%% file: main.tex
\documentclass[sigconf]{acmart}
\AtBeginDocument{%
  }

\setcopyright{acmlicensed}
\copyrightyear{2018}
\acmYear{2018}
\acmDOI{XXXXXXX.XXXXXXX}
\acmConference[Conference acronym 'XX]{Make sure to enter the correct
  conference title from your rights confirmation email}{June 03--05,
  2018}{Woodstock, NY}
\acmISBN{978-1-4503-XXXX-X/2018/06}

\usepackage{graphicx}
\usepackage{subcaption}   
\usepackage{amsmath}
\usepackage{booktabs}
\usepackage{amsfonts}
\usepackage{algorithm}
\usepackage{algorithmic}
\usepackage{multirow}
\usepackage[normalem]{ulem}     

\begin{document}

\title{PiXTime: A Model for Federated Time Series Forecasting with Heterogeneous Data across Nodes}


\author{Yiming Zhou}
\affiliation{%
  \institution{University of Science and Technology of China}
  \city{Hefei}
  \country{China}}
\email{zym2019@mail.ustc.edu.cn}

\author{Jiahao Wang}
\affiliation{%
  \institution{University of Science and Technology of China}
  \city{Hefei}
  \country{China}}
\email{jiahao.wang@mail.ustc.edu.cn}

\author{Mingyue Cheng}
\affiliation{%
  \institution{University of Science and Technology of China}
  \city{Hefei}
  \country{China}}
\email{mycheng@ustc.edu.cn}

\author{Hao Wang}
\affiliation{%
  \institution{University of Science and Technology of China}
  \city{Hefei}
  \country{China}}
\email{wanghao3@ustc.edu.cn}

\author{DeFu Lian}
\affiliation{%
  \institution{University of Science and Technology of China}
  \city{Hefei}
  \country{China}}
\email{liandefu@ustc.edu.cn}

\author{Enhong Chen}
\affiliation{%
  \institution{University of Science and Technology of China}
  \city{Hefei}
  \country{China}}
\email{cheneh@ustc.edu.cn}

\renewcommand{\shortauthors}{Zhou et al.}

\begin{abstract}
While collaborative forecasting on distributed time series is highly desirable, directly pooling localized datasets is often impractical due to data sharing constraints. Federated learning offers a promising alternative, yet conventional federated learning algorithms require homogeneous model architectures, which are incompatible with the structural discrepancies—specifically, unaligned temporal resolutions and mismatched variable channels—commonly observed across decentralized nodes. To bridge this gap, we introduce PiXTime, a novel Transformer-based framework designed to natively accommodate and leverage structurally heterogeneous temporal data. At its core, PiXTime adopts a parameter-decoupling architecture, strategically partitioning the model into localized personalized modules and a globally aggregated shared backbone. Specifically, node-specific local modules act as dimensional adapters, projecting raw sequences of diverse lengths into a unified representation space. Concurrently, a globally synchronized VE Table injects consistent categorical identities into the feature space, allowing the shared backbone to collaboratively learn and generalize representations across inconsistent variable distributions. Comprehensive evaluations on multiple benchmarks demonstrate that PiXTime achieves state-of-the-art performance in heterogeneous federated environments, while maintaining robust superiority in standard homogeneous and centralized forecasting settings.
\end{abstract}


\maketitle

\input{sections/introduction.tex}
\input{sections/related.tex}
\input{sections/pixtime.tex}

\input{sections/experiment.tex}
\input{sections/conclusion.tex}

\section*{GenAI Usage Disclosure}
Generative AI tools were used solely to assist with language editing and improving the logical flow of the manuscript. No generative AI was used in the research process, including idea development, methodology, experiment design, analysis, or coding. All intellectual contributions and scientific content are entirely the work of the authors, who take full responsibility for the accuracy and integrity of this paper.

\bibliographystyle{ACM-Reference-Format}
\bibliography{sample-base, sections/cite}

\end{document}

%% file: sections/introduction.tex
\section{Introduction}
\begin{figure}[t]
    \centering
    \includegraphics[width=\linewidth]{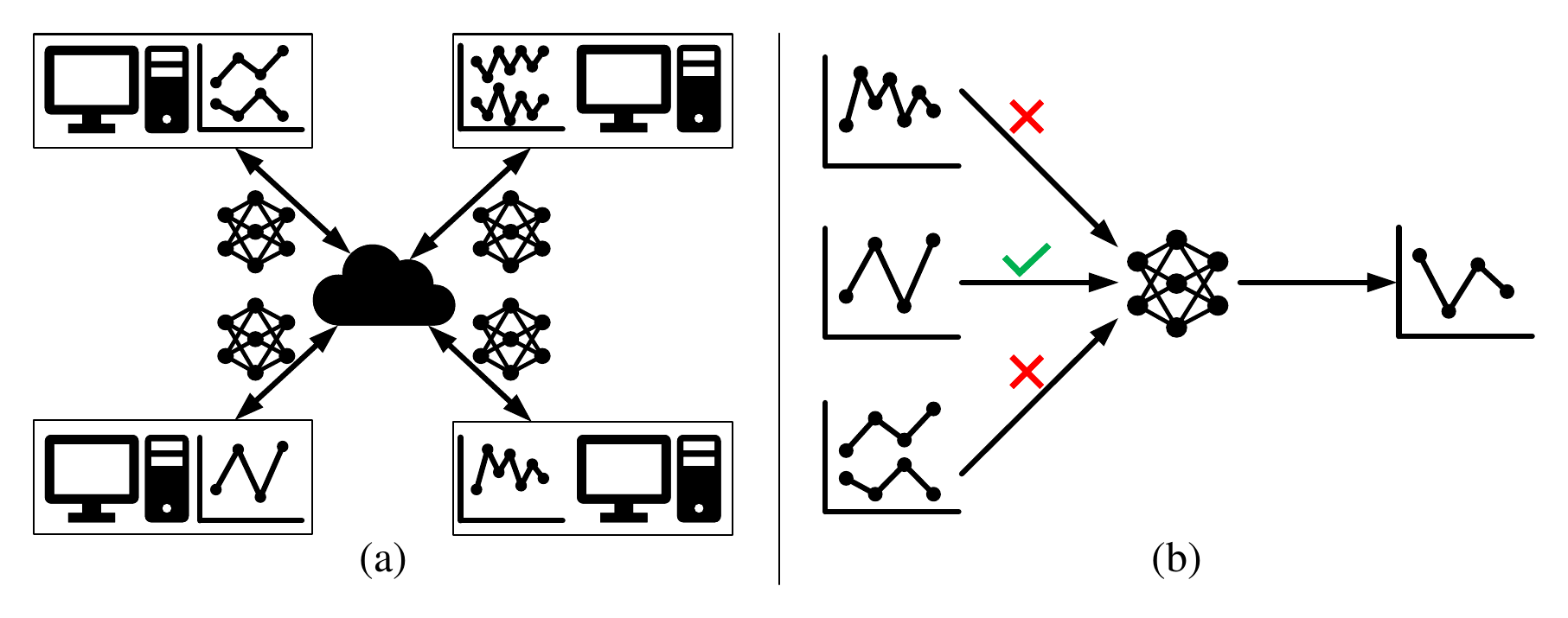}
    \caption{The dilemma of conventional federated learning algorithms when handling heterogeneous data structures. (a) Nodes with heterogeneous data collaboratively train a globally homogeneous model as required by the algorithms. (b) A strictly homogeneous model is generally incompatible with structurally heterogeneous data across nodes.}
    \label{fed_flow}
    \Description{None}
\end{figure}

Time series forecasting is a fundamental task that uses historical data of a variable to predict its future values. Many types of data in the real world can be recorded in a time series format, such as temperature, exchange rates, traffic flow, and electricity load \cite{Exchange,Informer}, making time series forecasting both essential and highly valuable. In practice, such valuable time series data are often recorded by multiple entities (such as hospitals, companies, and financial institutions), while tightening privacy policies and rising data security concerns are making it unrealistic to transmit raw data across them, resulting in numerous distributed data nodes \cite{CrossSiloFed,PerCrossSiloFed}. Meanwhile, the model trained by data from a single node often faces data distribution shifts caused by local data preferences, which in turn leads to a degeneration in their forecasting performance \cite{SCAFFOLD,VRSGD}. To meet data security requirements and reduce model degradation caused by data distribution shift, organizing federated learning \cite{FedAVG} among these nodes seems like a potential solution. However, due to the lack of unified data collection standards across entities, local datasets across nodes inevitably exhibit structural heterogeneity. These structurally heterogeneous data make it challenging for most state-of-the-art (SOTA) time series forecasting models to adapt to conventional federated optimization algorithms (e.g., FedAvg \cite{FedAVG}, FedProx \cite{FedProx}, pFedMe \cite{pFedMe}). This difficulty stems from a structural conflict shown in Fig.~\ref{fed_flow}: on the one hand, these federated algorithms require that the communicated models be homogeneous; otherwise, aggregation becomes impractical; on the other hand, the heterogeneous data structures across nodes typically require personalized, heterogeneous models for effective processing. Consequently, to deploy federated learning in such scenarios, the optimizing model must bridge the gap between locally heterogeneous data structures and globally homogeneous aggregation algorithms.

To address the aforementioned structural conflict, parameter decoupling \cite{FedPer,LGFEDAVG,FedRep,DFedPGP,ModelDecoupNonConvex} has emerged as a promising paradigm within the federated learning community. By carefully partitioning the model architecture, this paradigm bridges the gap between data heterogeneity and aggregation requirements without altering the conventional federated workflow. Under this paradigm, a complete model is explicitly divided into non-overlapping local and shared modules. Local modules are independently initialized and retained by each node to accommodate personalized data structures, and they are excluded from network communication. In contrast, shared modules are initialized by the central server or follow a unified configuration across nodes to guarantee global homogeneity, and they are communicated throughout the network during the federated training. This decoupling strategy naturally resolves the incompatibility between heterogeneous data and conventional aggregation algorithms: personalized local modules act as adapters that bridge structurally diverse raw data with a globally unified latent space, enabling the shared modules to process these aligned representations and leverage the robust generalization capability acquired through federated optimization. However, while the parameter decoupling paradigm is well-suited to this structural dilemma, the unique properties of time series data introduce novel complications.

In distributed time series data, structural heterogeneity primarily manifests in two ways: diverse temporal granularities and inconsistent variable categories. The first diverse temporal granularities challenge is caused by inconsistent sampling rates across nodes. Many recent SOTA Transformer-based forecasting models \cite{PatchTST,TimeXer,EAPformer,AdaPatch,FinCast} have achieved significant success by segmenting time series into patch tokens to capture temporal patterns. However, due to varying sampling rates, data segments covering the identical physical time interval consist of different numbers of data points across nodes. Consequently, a critical challenge is to design a local bridging mechanism that aligns these varying patch lengths into a unified representation dimension, thereby enabling the shared modules to absorb multi-granular temporal knowledge and achieve positive transfer that surpasses single-granular training. The second challenge originates from inconsistent variable categories, typically resulting from the inconsistent types of sensors deployed across distributed entities. In time series forecasting, incorporating historical data on relevant auxiliary variables often improves predictions of the target variable. Yet, in realistic scenarios, the available sets of auxiliary variables often exhibit significant heterogeneity, posing a fundamental challenge to effective prediction \cite{HyperIMTS}. To collaboratively leverage these heterogeneous features, it is necessary to establish a globally unified mechanism that consistently recognizes variable identities across nodes. Explicitly injecting this categorical awareness into the model's inputs is a crucial prerequisite for transferring variable-specific feature-processing capabilities, which ultimately improves forecasting performance in federated networks with highly inconsistent variable sets.



To address the aforementioned challenges, we propose the \textbf{P}atched target with \textbf{i}nverted au\textbf{X}iliary \textbf{Time}-series Transformer (\textbf{PiXTime}) model and its corresponding federated optimization framework. Built upon an encoder-decoder architecture, PiXTime follows the parameter-decoupling paradigm to segregate its internal components into personalized local modules and global shared modules. The personalized local modules act as tailored dimensional adapters that bridge node-specific temporal granularities with the globally unified latent space of the shared modules. Concurrently, to address inconsistent variable sets, we design a VE Table within the shared modules that establishes a globally synchronized category-identification mechanism. This explicitly enables consistent variable recognition and facilitates the cross-node transfer of variable-specific processing capabilities. The main contributions of this paper are summarized as follows:

\begin{itemize}
    \item We propose PiXTime, a novel Transformer-based model designed for federated time series forecasting. By embracing the parameter-decoupling paradigm, PiXTime segregates its architecture into personalized local modules and shared global modules, enabling collaborative learning while accommodating structurally heterogeneous data.
    
    \item We design specialized mechanisms to resolve structurally heterogeneous data: (i) the personalized local modules that act as dimensional adapters to bridge node-specific temporal granularities; and (ii) a globally synchronized VE Table providing a unified foundation to facilitate the cross-node transfer of feature-extraction capabilities.
    
    \item To effectively process the aligned sequences, we design an asymmetric Transformer encoder-decoder backbone tailored for PiXTime, where the encoder and decoder operate in representational spaces of distinct granularities. To connect these disparate representations, we introduce an abstract token as a granularity bridge to fuse cross-granular features, which also empowers PiXTime to leverage auxiliary information without requiring unaffordable computational overhead in resource-constrained federated settings.
    
    \item Extensive experiments on five benchmark datasets demonstrate that PiXTime outperforms SOTA baselines under various federated settings (both IID and Non-IID) and exhibits an exceptional capability to exploit structurally heterogeneous distributed data. Furthermore, PiXTime maintains highly competitive performance in the conventional single-node setting, confirming its remarkable versatility and robustness across diverse deployment scenarios.
\end{itemize}

%% file: sections/related.tex
\section{Related Work}\label{related_work}

\subsection{Parameter-Decoupling}

To practically implement structural heterogeneity and personalization in federated learning, parameter decoupling has emerged as a highly effective architectural strategy, which relies on an explicit layer-wise separation of the model. Pioneering works such as FedPer \cite{FedPer} and LG-FedAvg \cite{LGFEDAVG} proposed partitioning neural networks into shared global base layers and personalized local head layers. The feasibility of this paradigm was theoretically grounded by FedRep \cite{FedRep}, which proved the efficacy of sharing a global generic representation extractor while independently optimizing local classification heads. Further theoretical validations followed: Pillutla et al. \cite{ModelDecoupNonConvex} proved the convergence of this paradigm when optimizing non-convex objective functions, and Liu et al. \cite{DFedPGP} demonstrated its applicability in directed decentralized networks by integrating it with the Push-SUM protocol \cite{PS,AWPS}.

Subsequent advancements have substantially enriched this decoupling paradigm. Empirically, FedBABU \cite{FedBABU} validated the effectiveness of restricting server-side aggregation solely to the global body, keeping local heads strictly private for task-specific adaptation in classification scenarios. FedRoD \cite{FedRoD} introduced a decoupled objective by maintaining an additional personalized prediction head alongside the global head for each node, enabling the balance between global optimization and local personalized demands. Expanding on architectural techniques, FedDecomp \cite{FedDecomp} explored the application of low-rank parameter decomposition. Furthermore, the FedBone \cite{FedBone} algorithm verified the efficacy of parameter decoupling in complex models for healthcare scenarios, significantly contributing to its broader practical applicability. However, while parameter decoupling has achieved profound success in computer vision and various other domains, how to effectively design decoupled architectures to address the unique structural misalignments inherent in time series data remains an open problem.

\subsection{Time Series Forecasting}


The ``Patching'' technique has recently emerged as a widely adopted paradigm in Transformer-based time series forecasting. Departing from conventional pointwise tokenization, PatchTST \cite{PatchTST} segments continuous 1D time series into localized subsequences (patches), which drastically reduces sequence length while preserving local semantic context. This paradigm has since been extended by several innovative variants: AdaPatch \cite{AdaPatch} adopts an adaptive scheme for patch-level encoding and normalization to address intra-instance distributional shifts, while EAPformer \cite{EAPformer} proposes an entropy-aware patching method to dynamically segment time series for differentiated assessments of historical patterns. Furthermore, TimeFM \cite{TimeFM} investigates the potential of patch-based, decoder-only architectures to establish foundation models in the time series domain, while FinCast \cite{FinCast} further explores the application of Mixture-of-Experts (MoE) architectures for such models. More recently, TimeXer \cite{TimeXer} introduced a hybrid tokenization strategy that employs an abstract token to bridge different representational granularities. While PiXTime shares a similar motivation for using abstract tokens, its core architecture differs significantly in the design of the variable-wise representation extractor. Specifically, TimeXer adopts a linear extractor that operates on fixed-length inputs, which limits its ability to handle varying sequence lengths and to transfer variable-specific extraction capabilities across structurally heterogeneous nodes. In contrast, PiXTime employs a length-agnostic architecture that supports varying-length sequences, facilitating cross-node knowledge transfer for structurally heterogeneous nodes.

How to effectively coordinate distributed nodes for federated time-series forecasting is attracting rapidly growing interest. Early works such as FedTime \cite{FedTime} and Time-FFM \cite{TimeFFM} explored leveraging pretrained Large Language Models (LLMs) for this task, with FedTime fine-tuning the pretrained backbone and Time-FFM keeping it frozen. However, these LLM-based approaches are computationally prohibitive for resource-constrained edge nodes. To address data distribution diversity, Fed-TREND \cite{FedTREND} generates informative synthetic data as auxiliary knowledge carriers; FFTS \cite{FFTS} introduces an encoder-only model with a MoE design and a tailored optimization framework; and FeDal \cite{FeDal} explicitly mitigates knowledge bias across distributed datasets through domain and global bias elimination. Furthermore, several efforts have investigated federated time-series forecasting in specific domains, such as medicine \cite{Ali} and climate \cite{WeatherLLM}. Tang et al. \cite{Tang} theoretically examined how introducing a synthetic data generator can resolve temporal heterogeneity, yet their framework lacks extensive empirical validation. More recently, ProtoPFL \cite{ProtoPFL} combined prototype-based personalized federated learning to deliver a differential-privacy-preserving training framework, but left the critical challenge of structural data heterogeneity undiscussed. FedRMamba \cite{FedRMamba} tackles the federated scenarios' variable heterogeneity problem in the frequency domain and shares a frequency-aware Mamba that mines low-frequency dependencies among nodes, thereby transferring feature extraction capabilities. However, compared with PiXTime, FedRMamba requires all nodes to share the same lookback window size, which makes it lack the ability to transfer multi-granularity temporal feature extraction. To the best of our knowledge, we are the first to simultaneously address both variable and temporal granularity heterogeneities for federated time-series forecasting.

%% file: sections/pixtime.tex
\begin{figure*}[t]
    \centering
    \includegraphics[width=\linewidth]{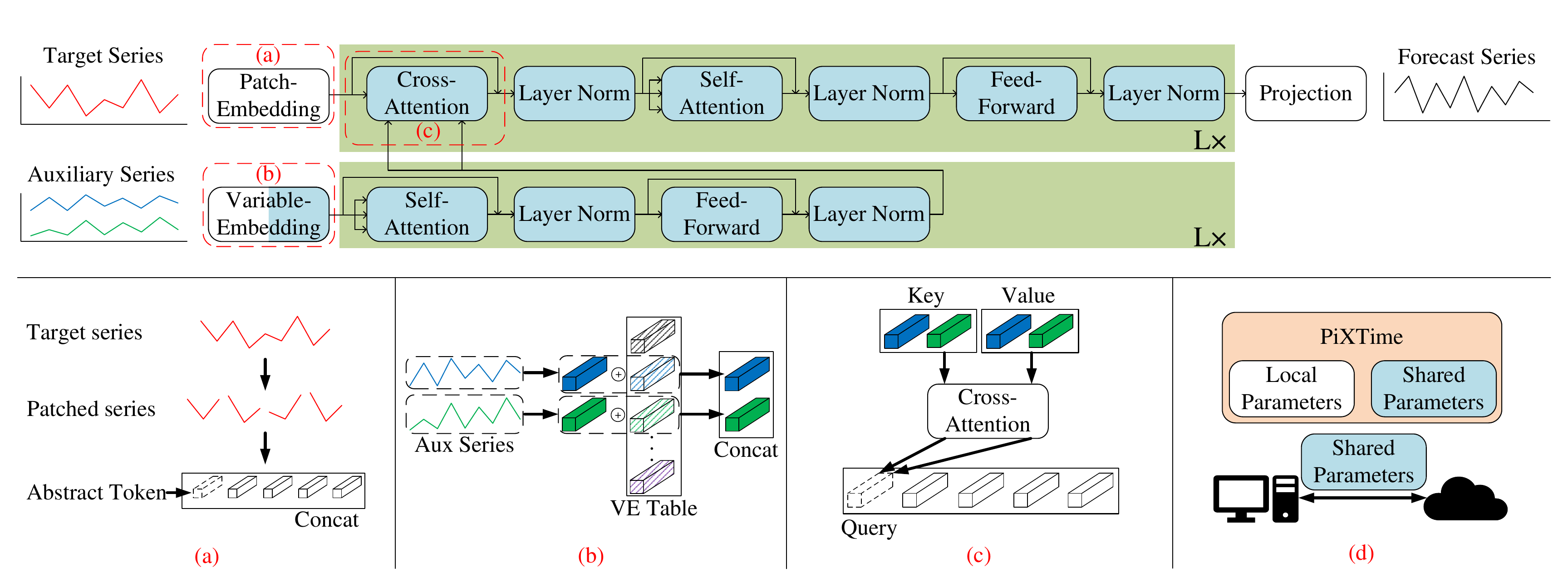}
    \caption{The architecture of PiXTime. (a) The target series is sliced temporally and mapped into a token sequence, then an abstract token is concatenated for subsequent processing. (b) Multiple auxiliary series are mapped variable-wise into a token sequence, then corresponding variable embeddings are retrieved and added before further processing. (c) Representations from auxiliary series are transferred into the abstract token exclusively through cross-attention. (d) The blued modules of PiXTime are shared by federated learning, while the rest are kept local to each node.}
    \label{pixtime_flow}
    \Description{None}
\end{figure*}

\section{PiXTime}

%

\textbf{Problem Setting.} Consider a network of $N$ nodes collaborating via a centralized server and a specific federated learning algorithm to optimize their models. Take the multivariate-to-univariate forecasting task as an example, in which all nodes aim to predict a common target variable and are equipped with heterogeneous sets of auxiliary variables to support their predictions. For a node $i$ in the network, its PiXTime takes as input a target series $\textbf{x}_{i} = \{ x_1, x_2, \cdots , x_{T_i} \}\in \mathbb{R}^{T_i} $ and multiple auxiliary series $\textbf{Z}_{i} = \{ \textbf{z}_{i}^{(1)}, \textbf{z}_{i}^{(2)}, \cdots , \textbf{z}_{i}^{(C_i)} \}\in \mathbb{R}^{T_i\times C_i}$, then outputs a prediction $\textbf{x}_{i}^{pre} = \{ x_{T_i+1}, x_{T_i+2}, \cdots , x_{T_i+S_i} \}\in \mathbb{R}^{S_i}$, where $T_i$ is the look-back window length, $C_i$ is the number of auxiliary variables, and $S_i$ is the prediction length. By comparing the output $\textbf{x}_{i}^{pre}$ with the ground truth $\textbf{y}_{i} = \{ y_{T_i+1}, y_{T_i+2}, \cdots , y_{T_i+S_i} \}\in \mathbb{R}^{S_i} $, node $i$ can optimize its PiXTime parameters $\theta_i $, and subsequently share a subset of these parameters with server to enable federated learning. We assume that the sequences across all nodes span the same physical time intervals, while allowing their lengths to vary due to different temporal granularities (e.g., client-specific $T_i$ and $S_i$). Additionally, we allow personalized auxiliary series matrices across nodes, with node-specific numbers of variables $C_i$, and node-specific variable categories for each auxiliary series in $\textbf{Z}_i$, which are stored in $\textbf{Z}_{i}^{cat}$.

\subsection{Model Workflow}
We use the multivariate-to-univariate (M2U) forecasting task as an example to detail the model workflow of PiXTime. For the multivariate-to-multivariate (M2M) and univariate-to-univariate (U2U) forecasting, we briefly outline their analogous processes at the end of this subsection. As shown in Fig.~\ref{pixtime_flow}, the architecture of PiXTime is a variant of Transformer that incorporates multiple specialized modifications tailored for time series forecasting. 


\noindent
\textbf{Patch Embedding.} The Patch Embedding module of PiXTime on node $i$ takes the target series $x_i$ of length $T_i$ as input. It first segments the sequence into a series of patches of length $PL_i$ via a parameter-free slicing operation. Subsequently, a linear projection layer maps this patch series into patch tokens, which are then concatenated with an abstract token and forwarded to the downstream components. We formalize this processing as follows:
\begin{equation}\label{PE}
\begin{aligned}
&\left \{ \textbf{p}_{i}^{(1)}, \textbf{p}_{i}^{(2)}, \cdots , \textbf{p}_{i}^{(M_i)} \right \} =PatchSplitter\left ( \textbf{x}_i \right ),\\
&\textbf{P}_i  =PatchLinear\left ( \textbf{p}_{i}^{(1)}, \textbf{p}_{i}^{(2)}, \cdots , \textbf{p}_{i}^{(M_i)}\right ),\\
&\left [ \textbf{a}_i,\textbf{P}_i \right ]  =Concat\left ( \textbf{a}_i,\textbf{P}_i\right ),
\end{aligned}
\end{equation}
\noindent
where $M_i$ is the number of patches after slicing. The slicing stride can be flexibly set to allow overlapping or non-overlapping patches; for a non-overlapping setting, $M_i=T_i / PL_i$. $PatchSplitter$ splits $\textbf{x}_{i}$ into patches $\left \{ \textbf{p}_{i}^{(1)}, \textbf{p}_{i}^{(2)}, \cdots , \textbf{p}_{i}^{(M_i)} \right \}$, each in $\mathbb{R}^{PL_i}$, with no learnable parameters involved. $PatchLinear$ maps these patches into a token sequence $\textbf{P}_i \in \mathbb{R}^{D\times M_i}$ via a linear layer. $Concat$ adds a learnable abstract token $\textbf{a}_i \in \mathbb{R}^{D}$ at the beginning of the token sequence to capture its variable-wise representation.

The integration of the abstract token addresses the asymmetric representation strategies applied to the target and auxiliary variables in PiXTime. Because the target variable is the primary forecasting objective, we employ fine-grained patch tokens to capture its detailed temporal patterns. Conversely, recognizing that auxiliary variables contribute unequally to the prediction, we treat each auxiliary sequence as a single token to extract variable-wise representations. This forces the model to learn the inter-variable dependencies. However, this asymmetric design inherently creates a granularity mismatch between the encoder (operating in a variable-wise space) and the decoder (operating in a patch-wise space). To bridge this semantic gap, inspired by~\cite{TimeXer}, we append an abstract token to the target's patch token sequence. This abstract token functions to aggregate a variable-level representation of the target sequence, serving as a vital bridge in subsequent cross-attention modules to align the patch-based target features with the variable-based auxiliary features.


\noindent
\textbf{Variable Embedding.} To capture the aforementioned inter-variable dependencies of auxiliary, the Variable Embedding on node $i$ processes the auxiliary series set $Z_i$ by independently projecting the sequences of its $C_i$ variables into $C_i$ variable tokens. Subsequently, based on each variable's category, a corresponding $D$-dimensional variable embedding is retrieved from the VE Table and added to the respective token. This variable embedding injects explicit categorical identity into each token, enabling downstream modules to distinguish different variable categories and thereby enhancing the extraction of inter-variable dependencies. We formalize this processing as follows:
\begin{equation}\label{VE}
\begin{aligned}
& \textbf{V}_i =VELinear\left ( \textbf{Z}_i \right ),\\
&\textbf{V}_i^{aux}  = \textbf{V}_i + VETable\left ( \textbf{Z}_i^{cat} \right ).
\end{aligned}
\end{equation}
$VELinear$ maps the input auxiliary series set $\textbf{Z}_{i}\in \mathbb{R}^{T_i\times C_i}$ into a variable token sequence $\textbf{V}_i\in \mathbb{R}^{D\times C_i}$ via a variable-wise projection. The $VETable$ retrieves a $D$-dimensional learnable embedding for each variable token based on its specific category in $\textbf{Z}_i^{cat}$ (the initialization and server-node synchronization mechanisms of $VETable$ are detailed in subsequent sections). These retrieved variable embeddings are then added to their corresponding tokens in $\textbf{V}_i$, yielding the output $\textbf{V}_i^{aux}\in \mathbb{R}^{D\times C_i}$.


\noindent
\textbf{Auxiliary Encoder.} This module is a variant of the Transformer encoder, designed to process the variable token sequence $\textbf{V}_i^{aux}$ generated by the Variable Embedding. Its primary function is to extract the variable-wise representations of auxiliary variables, which subsequently assist the decoder as the key and value matrices in forecasting the target series. This process is formalized as follows:
\begin{equation}\label{AE}
\begin{aligned}
&\textbf{V}_{i,0}^{aux}= \textbf{V}_i^{aux};\\
&\textbf{V} _{i,l+1}^{aux} = Aux\text{-}Encoder\left ( \textbf{V}_{i,l}^{aux} \right ) ,\ l=0,1,\cdots,L-1,
\end{aligned}
\end{equation}
where $L$ is the number of layers in PiXTime’s encoder-decoder module. The $Aux\text{-}Encoder$ applies attention-based weighting to the input $\textbf{V}_i^{aux}$ to extract variable-wise representations, outputting $\textbf{V} _{i,L}^{aux}\in \mathbb{R}^{D\times C_i}$ to support PiXTime’s subsequent forecasting.


\noindent
\textbf{Target Decoder.} This module primarily consists of a customized cross-attention block and a standard self-attention block, receiving the target patch tokens $\textbf{P}_i$ and the abstract token $\textbf{a}_i$ from Patch Embedding, alongside the encoded auxiliary features $\textbf{V} _{i,L}^{aux}$ from Auxiliary Encoder. As previously stated, a fundamental granularity gap exists: $\textbf{P}_i$ operates in a fine-grained, patch-wise temporal space, whereas $\textbf{V} _{i,L}^{aux}$ operates in a coarse-grained, variable-wise semantic space. To bridge this disparity, PiXTime utilizes the abstract token to aggregate the variable representation of the target sequence. Operating within the same variable-level space as the auxiliary features, the abstract token is exclusively assigned as the query in the cross-attention module to extract relevant auxiliary knowledge. Subsequently, the abstract token distributes the auxiliary knowledge to target patches by a self-attention component to guide the target variable's forecasting. We formalize this as follows:
\begin{align}
&\left [ \textbf{a}_{i,0},\textbf{P}_{i,0} \right ] =\left [ \textbf{a}_i,\textbf{P}_i \right ];\notag\\
&\ \textbf{a}_{i,l}^{cro} =LN\left ( \textbf{a}_{i,l} + Cross\text{-}Att\left ( \textbf{a}_{i,l}, \textbf{V} _{i,L}^{aux}, \textbf{V} _{i,L}^{aux} \right )  \right ) ,\notag\\
&\left [ \overline{\textbf{a}}_{i,l}^{cro},\overline{\textbf{P}}_{i,l} \right ]  =LN\left ( \left [ \textbf{a}_{i,l}^{cro},\textbf{P}_{i,l} \right ]  + Self\text{-}Att\left ( \left [ \textbf{a}_{i,l}^{cro},\textbf{P}_{i,l} \right ]  \right )  \right ) ,\label{TD}\\
&\left [ \textbf{a}_{i,l+1},\textbf{P}_{i,l+1} \right ]  =LN\left ( \left [ \overline{\textbf{a}}_{i,l}^{cro},\overline{\textbf{P}}_{i,l} \right ]   + FFN\left ( \left [ \overline{\textbf{a}}_{i,l}^{cro},\overline{\textbf{P}}_{i,l} \right ]   \right )  \right ),\notag\\
&\quad  \quad  \quad \quad \quad \quad \quad \quad \quad \quad \quad \quad \quad\quad\quad\quad  l=0,1,\cdots,L-1.\notag
\end{align}

\noindent
For each layer $l$ within the $L$-layer Target Decoder, it first weights the abstract token $\textbf{a}_{i,l}$ with $\textbf{V}_{i,L}^{aux} $ from the Auxiliary Encoder through a cross-attention block to obtain $\textbf{a}_{i,l}^{cro}$. Then, it distributes the auxiliary knowledge in $\textbf{a}_{i,l}^{cro}$ to the target patch sequence $\overline{\textbf{P}}_{i,l}$ and aggregates the variable-wise representation of the target variable into the abstract token $\overline{\textbf{a}}_{i,l}^{cro}$ through a self-attention block. Finally, the sequence is processed by a Feed-Forward Network (FFN) before proceeding to the next layer.


\noindent
\textbf{Projection Head.} This module is responsible for processing the output of the Target Decoder to generate PiXTime's final prediction. Crucially, the final layer of the Target Decoder outputs both the abstract token $\textbf{a}_{i,L}$ and the patch tokens $\textbf{P}_{i,L}$, which inherently encapsulate variable-level and patch-level representations, respectively. Because the forecasting task needs reconstructing fine-grained temporal sequences, the final prediction is exclusively derived from the temporal patch tokens $\textbf{P}_{i,L}$, discarding the abstract token.
\begin{equation}\label{PH}
\textbf{x}_i^{pre} = Projection\left ( \textbf{P}_{i,L} \right ),
\end{equation}
\noindent
where $\textbf{x}_i^{pre} \in \mathbb{R}^{S_i}$ represents the final forecasted sequence obtained by the PiXTime model on node $i$.

\noindent
\textbf{M2M and U2U.} For the M2M task, since all variables are targets, the Patch Embedding module processes the entire multivariate matrix and appends an independent abstract token to each variable's patch tokens. Subsequently, each abstract token acts as an independent query to perform cross-attention with variable tokens from the Auxiliary Encoder. For the U2U task, to maintain the architectural computational flow, the single target sequence is simply duplicated and fed into the Auxiliary Encoder as a self-auxiliary variable.

\subsection{Federated Optimization}

%





We elaborate on the federated optimization of PiXTime using the FedAvg framework; the extension to FedProx (as used in our experiments) can be derived similarly.

\subsubsection{Optimization Algorithm}
\begin{figure}[t]
    \centering
    \includegraphics[width=0.6\linewidth]{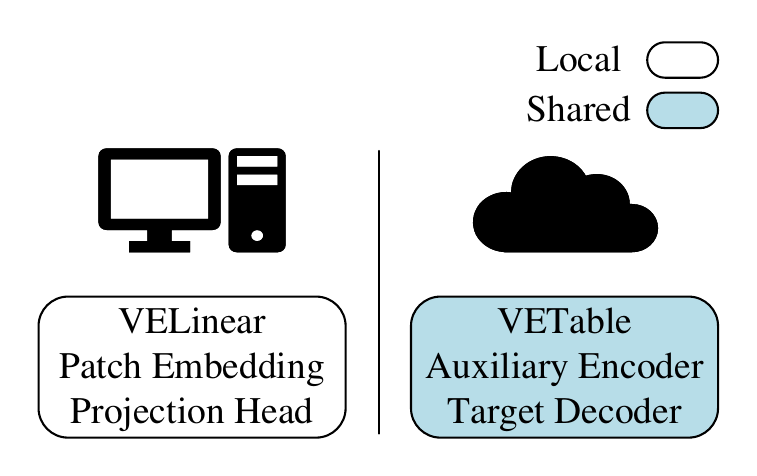}
    \caption{Decoupling of PiXTime: The division of local modules of nodes and shared modules for the server.}
    \label{fed_init}
    \Description{None}
\end{figure}

In FedAvg \cite{FedAVG}, the central server requires architectural homogeneity across the model parameters collected from nodes for aggregation (step 13 in Algorithm~\ref{pixtime_fedavg}). By examining the modules of PiXTime, we observe that the VE Table, Auxiliary Encoder, and Target Decoder across different nodes are identically designed based on the token dimension $D$. Consequently, these modules are aggregatable by the server. Conversely, modules such as the VELinear, Patch Embedding, and Projection Head are structurally tied to node-specific configurations, rendering them normally unaggregatable. To resolve this structural conflict, as illustrated in Fig.~\ref{fed_init}, we explicitly decouple PiXTime's modules into local and shared. The local modules are independently initialized and kept by each node, whereas the shared modules are maintained and aggregated by the centralized server.

\begin{algorithm}[t]
\caption{Optimization of PiXTime (based on FedAvg)}
\label{pixtime_fedavg}
\begin{algorithmic}[1]
\renewcommand{\algorithmicrequire}{\textbf{Init:}}
\renewcommand{\algorithmicensure}{\textbf{Output:}}
\REQUIRE Shared modules' parameters $\theta_s^{(0)}$ on server; Local personalized modules' parameters $\theta_{l,i}^{(0)}$ for each node $i\in \{1,\dots,N\}$; Learning rate $\gamma$, rounds $T$, local epochs $E$.
\ENSURE $\theta_s^{(T)}$ and $\theta_{l,i}^{(T)}$ for each node $i \in \{1,\dots,N\}$.

\FOR{$t = 0, \dots, T-1$}
    \STATE The server selects a node subset $S^{(t)} \subseteq \{1, \dots, N\}$.
    \STATE The server broadcasts $\theta_s^{(t)}$ to each node $i \in S^{(t)}$.
    
    \FOR{each node $i \in S^{(t)}$ \textbf{in parallel}}
        \STATE Construct full model: $\theta_i^{(t,0)} \leftarrow [\theta_s^{(t)}, \theta_{l,i}^{(t)}]$.
        \FOR{epoch $e = 0, \dots, E-1$}
            \STATE Sample mini-batch $\xi_i^{(t,e)}$ from local dataset $\mathcal{D}_i$.
            \STATE $\theta_i^{(t,e+1)} \leftarrow \theta_i^{(t,e)} - \gamma \nabla F\left( \theta_i^{(t,e)} ; \xi_i^{(t,e)} \right)$.
        \ENDFOR
        \STATE Split $\theta_i^{(t,E)}$ into updated $[\theta_{s,i}^{(t+1)}, \theta_{l,i}^{(t+1)}]$.
        \STATE Send $\theta_{s,i}^{(t+1)}$ to the server.
    \ENDFOR
    
    \STATE The server aggregates: 
    $\theta_{s}^{(t+1)} \leftarrow \frac{1}{|S^{(t)}|}\sum_{i \in S^{(t)}}  \theta_{s,i}^{(t+1)}$.
    \STATE Unselected nodes ($i \notin S^{(t)}$): $\theta_{l,i}^{(t+1)} \leftarrow \theta_{l,i}^{(t)}$.
\ENDFOR
\end{algorithmic}
\end{algorithm}


Algorithm~\ref{pixtime_fedavg} provides a comprehensive workflow of the federated optimization process for PiXTime. During initialization, the centralized server initializes the global shared modules with parameters $\theta_s^{(0)}$, while each node independently initializes its personalized local modules with parameters $\theta_{l,i}^{(0)}$. In each round, the server activates a subset of nodes and broadcasts the current shared modules to them. Upon receiving the shared modules, the active nodes assemble them with their respective local modules to construct a complete PiXTime model. This assembled model is then optimized using the node's local dataset, with the Mean Squared Error uniformly adopted as the loss function $F$ across all nodes. After completing the local optimization steps, each active node divides the updated PiXTime model back into its shared and local modules. The updated local modules are stored strictly on the node, while the updated shared modules are transmitted back to the server. Finally, the server aggregates the parameters of the shared modules collected to conclude the current round. 

\noindent \textbf{Remark on Convergence:} We refer to the non-convex analysis of decoupled federated learning established in \cite{ModelDecoupNonConvex} for the convergence of Algorithm \ref{pixtime_fedavg}. By mapping our local personalized parameters $\theta_{l,i}$ to the local variables $v_i$ and the shared parameters $\theta_s$ to the global variables $u$ in \cite{ModelDecoupNonConvex}, Algorithm \ref{pixtime_fedavg} shares the identical optimization framework as their FedSim algorithm. Under the standard optimization assumptions outlined in \cite{ModelDecoupNonConvex} (e.g., L-smoothness and bounded variance), the convergence guarantees established for FedSim directly apply to our algorithmic instantiation.

\subsubsection{Resource Overhead} In federated learning scenarios, the computational and communication resources of edge nodes are often limited. Therefore, we examine the hardware requirements of implementing Algorithm~\ref{pixtime_fedavg} to optimize PiXTime. We run a profiling experiment to measure the communication payload, peak memory footprint, and computational overhead required for the local optimization of PiXTime, while investigating the reduction in overhead achieved by the Target Decoder design, in which only the abstract token participates in cross-attention.

The experimental settings and profiling results are presented in Table~\ref{tab:pixtime_specs}, where ``Ori'' denotes the original PiXTime and ``Abs'' denotes PiXTime without the abstract token and its associated processing steps completely removed. More detail, in ``Abs'' variant, the critical cross-attention mechanism bridging the encoder and decoder (originally formulated in Eq.~\eqref{TD}) is forced to directly use the fine-grained patch tokens as queries:
\begin{equation}\label{Abs}
\textbf{P}_{i,l}^{cro} = LN\left ( \textbf{P}_{i,l} + Cross\text{-}Att\left ( \textbf{P}_{i,l}, \textbf{V} _{i,L}^{aux}, \textbf{V} _{i,L}^{aux} \right ) \right ).
\end{equation}
 
As presented in Table~\ref{tab:pixtime_specs}, the results for the ``Ori'' setting indicate that the local updating of PiXTime is highly efficient. A maximum of 224.90 MB peak memory is required per node, while the computational costs for the forward and backward passes are merely 7.43B and 14.81B FLOPs, respectively. Additionally, the communication load for a single transmission with the server is only 15.64 MB. These metrics demonstrate that PiXTime’s resource overhead is well acceptable for edge nodes. Furthermore, it is well-known that the computational complexity of cross-attention scales quadratically with the number of query tokens. The introduction of the abstract token in the original PiXTime restricts the number of queries to exactly one per target variable. Comparing the overheads of ``Ori'' and ``Abs'' in Table 1, the abstract token mechanism directly reduces the peak memory footprint by 10.77\%, forward computational cost by 30.56\%, and backward computational cost by 30.66\%. This reduction is critically beneficial for resource-constrained edge devices. Finally, the ``Ori'' results verify that PiXTime’s parameter-decoupling design intrinsically reduces communication costs. By keeping local modules, the number of communicated parameters per node is reduced by 22.26\%, dropping from the full 10.06M model parameters to 7.82M shared parameters. In conclusion, the resource overhead of PiXTime is highly manageable for federated deployment.

\begin{table}[t]
\centering
\caption{Resource Overhead to Optimize PiXTime.}
\label{tab:pixtime_specs}
\begin{tabular}{lccc}
\toprule
Metric & Ori & Abs & Unit \\
\midrule
Total Model's Parameters & 10.06 & 10.04 & Million \\
Shared Modules' Parameters & 7.82 & 7.82& Million \\
Upload Bandwidth (bf16) & 15.64 & 15.64& MB  \\
Max Memory Footprint & 224.90 & 252.07 & MB  \\
Forward Pass FLOPs & 7.43 & 10.70 & Billion \\
Backward Pass FLOPs & 14.81 & 21.36 & Billion \\
\bottomrule
\end{tabular}
\raggedright
\\
\footnotesize{Note: Evaluated on a 36-variable M2M forecasting task (batch size 1). Architecture settings: look-back window 720, patch length 24, prediction length 120, model dimension 512, FFN dimension 2048, and 1 layer.}
\end{table}

\subsubsection{Structural Heterogeneity} As detailed in Algorithm~\ref{pixtime_fedavg}, each node trains on its local dataset. In realistic federated scenarios, these distributed datasets inevitably exhibit structural heterogeneity. For time series forecasting, its structural heterogeneity primarily manifests across nodes as: (1) misaligned temporal granularities caused by diverse sampling rates (e.g., varying look-back windows and prediction lengths), and (2) inconsistent auxiliary variable sets arising from various channel combinations (e.g., varying components within the set). To handle these structural heterogeneities during federated optimization, PiXTime leverages personalized local modules and the global shared VE Table, respectively.

Specifically, PiXTime's personalized local modules are designed to resolve the misaligned temporal granularities. To ensure the shared modules operate on a unified time scale across nodes, PiXTime requires all nodes to segment their respective inputs and outputs into consistent physical time intervals. However, due to misaligned temporal granularities caused by diverse sampling rates, identical physical time intervals correspond to different sequence lengths across nodes. To bridge these varying sequence lengths with the fixed dimension of the shared model, we construct the node-specific local modules illustrated in Fig.~\ref{fed_init}. By employing these personalized modules, PiXTime's shared module can seamlessly perform federated optimization across nodes with diverse sampling rates, aggregating multi-granular knowledge to facilitate positive transfer and enhance overall forecasting performance.

\begin{table*}[t]
\centering
\caption{The M2M forecasting average results under non-federated settings. Best results are in {\textbf{bold}}, second best are \uline{underlined}.}
\label{single}
\begin{tabular}{ccccccccccccc}
\toprule
Model  & \multicolumn{2}{c}{PiXTime (Ours)} & \multicolumn{2}{c}{DLinear} & \multicolumn{2}{c}{iTransformer} & \multicolumn{2}{c}{PatchTST} & \multicolumn{2}{c}{TimeXer} & \multicolumn{2}{c}{FFTS}\\
\cmidrule(lr){1-1}\cmidrule(lr){2-3} \cmidrule(lr){4-5} \cmidrule(lr){6-7} \cmidrule(lr){8-9} \cmidrule(lr){10-11} \cmidrule(lr){12-13} 
Metric & MSE & MAE & MSE & MAE & MSE & MAE & MSE & MAE & MSE & MAE &MSE & MAE\\
\midrule
ETT & 0.377 & \textbf{0.392} & 0.447 & 0.447 & 0.389 & 0.403 & 0.381 & 0.401 & \textbf{0.374} & \uline{0.392} & \uline{0.376} & 0.395 \\
\midrule
Electricity & \textbf{0.186} & \uline{0.282} & 0.225 & 0.319 & \uline{0.190} & \textbf{0.277} & 0.208 & 0.297 & 0.193 & 0.287 & 0.210 & 0.296 \\
\midrule
Traffic & \textbf{0.486} & \textbf{0.319} & 0.673 & 0.419 & \uline{0.488} & \uline{0.327} & 0.537 & 0.348 & 0.543 & 0.358 & 0.558 & 0.358 \\
\midrule
Exchange & 0.371 & 0.411 & \textbf{0.355} & 0.417 & 0.415 & 0.442 & 0.376 & \uline{0.409} & 0.407 & 0.427 & \uline{0.364} & \textbf{0.404} \\
\midrule
Weather & \textbf{0.245} & \textbf{0.273} & 0.264 & 0.315 & 0.260 & 0.281 & 0.256 & 0.278 & \uline{0.245} & \uline{0.273} & 0.257 & 0.279 \\
\bottomrule
\end{tabular}
\end{table*}

The VE Table is designed to handle the inconsistent auxiliary variable set across nodes, with the objective of enabling PiXTime to explicitly identify and differentiate diverse variable categories. To achieve this, the VE Table is designed as a shared module, initialized and maintained by the central server. Structurally, it operates as a dictionary-like parameter class, where each key represents a specific variable category, and the corresponding value is a $D$-dimensional learnable embedding. During initialization, the server creates a unique learnable embedding for every known variable category to construct the global VE Table. This table is subsequently broadcast to active nodes as one of the shared modules. During local updates, if an active node encounters a variable category absent from the current VE Table, it initializes a temporary embedding for this category to ensure that local training proceeds. The node then reports this new category (without the temporary embedding) to the server in the subsequent upload, allowing the server to initialize a globally unique embedding for this category and append it to the global VE Table, ensuring consistent recognition of this variable in subsequent training. After the active nodes complete their local updates, their respective VE Tables are uploaded and aggregated at the server. This continuous federated synchronization guarantees that variable embeddings remain globally consistent, enabling PiXTime to uniformly identify variables across nodes with diverse auxiliary variable sets. Consequently, this alignment enables a node to transfer its representation extraction capability for specific variables to others that share the same variables, enhancing the performance of the federated network in handling variable heterogeneity.

%% file: sections/experiment.tex
\section{Experiments}
To comprehensively verify the effectiveness and generality of PiXTime, we first compare it with SOTA time series forecasting models under conventional non-federated settings, as most of the baselines are designed for such scenarios. We then evaluated the forecasting performance of PiXTime under the federated setting and validated through ablation studies the effectiveness of its core designs.


\noindent
\textbf{Datasets.} We include five popular datasets: Electricity, Traffic, Weather, Exchange \cite{Exchange,Autoformer}, and the ETT datasets (including ETTh1, ETTh2, ETTm1, ETTm2) \cite{Informer}. These datasets are widely used in long-term forecasting benchmarks and are publicly available. They not only contain multiple variables, with Electricity and Traffic each including hundreds of variables, but also exhibit diverse sampling rates, from minutes to daily. These properties make them suitable for PiXTime’s experiments.


\noindent
\textbf{Baselines.} We chose five SOTA time series forecasting models: DLinear \cite{DLinear}, iTransformer \cite{iTrans}, PatchTST \cite{PatchTST}, TimeXer \cite{TimeXer}, and FFTS \cite{FFTS}. Among these baselines, DLinear operates as a purely linear model, whereas the others are Transformer-based architectures. Specifically, iTransformer processes each variable's entire series (including the target) as an individual variable token. PatchTST segments every variable sequence into patches and sequentially inputs these patch tokens into the model. TimeXer adopts a hybrid approach, feeding the abstract token along with the patch tokens into the projection head for the final prediction. Since the four models above were originally designed for traditional non-federated forecasting, we also include FFTS, a specialized Transformer-encoder-only architecture developed for federated time series forecasting, to provide a comprehensive comparison with PiXTime.

\noindent
\textbf{Remark to Baseline:} Although FedRMamba \cite{FedRMamba} addresses a problem similar to PiXTime (but without consideration of heterogeneous temporal granularities) and should be included as a baseline, as of CIKM's deadline (May 23), the official repository of FedRMamba has not yet released its model implementation. Therefore, we have to forgo including it as a baseline.


\noindent
\textbf{Implementation Details.} Our experiments follow an open-source benchmark \cite{Code}, including the implementations of baseline models, evaluation metrics, and hyperparameter settings. Unless otherwise stated, all experiments use the widely-used settings, with a look-back window of 96, prediction lengths of \{96, 192, 336, 720\}, a patch length of 16, and non-overlapped patching following PatchTST \cite{PatchTST}; For the non-federated experiments, we use the Adam optimizer; For the federated experiments, we use Algorithm 1 to optimize the models, with network scale $N=8$, local epochs $E=5$, and a fraction of activated nodes 50\%. We report results using the Mean Squared Error (MSE) and the Mean Absolute Error (MAE) metrics, with lower values indicating better performance for both. Detailed hyperparameter configurations are available in our repository.

\begin{table*}[t]
\centering
\caption{The M2M forecasting average results under federated settings. Best results are in {\textbf{bold}}, second best are \uline{underlined}.}
\label{fed}
\begin{tabular}{c|ccccccccccccc}
\toprule
\multicolumn{2}{c}{Model}  & \multicolumn{2}{c}{PiXTime (Ours)} & \multicolumn{2}{c}{DLinear} & \multicolumn{2}{c}{iTransformer} & \multicolumn{2}{c}{PatchTST} & \multicolumn{2}{c}{TimeXer} &\multicolumn{2}{c}{FFTS}\\
\cmidrule(lr){1-2}\cmidrule(lr){3-4} \cmidrule(lr){5-6} \cmidrule(lr){7-8} \cmidrule(lr){9-10} \cmidrule(lr){11-12} \cmidrule(lr){13-14} 
\multicolumn{2}{c}{Metric} & MSE & MAE & MSE & MAE & MSE & MAE & MSE & MAE & MSE & MAE& MSE & MAE \\
\midrule
\multirow{2}{*}{ETT} & IID   & \textbf{0.367} &	\textbf{0.388} &	0.441 &	0.442 &	0.388 &	0.403 &	0.390 &	0.406 &	\uline{0.379} &	\uline{0.397} &	0.386 &	0.401    \\
& non-IID  & \textbf{0.386} &	\textbf{0.397} &	0.452 &	0.452 &	0.401 &	0.409 &	0.424 &	0.418 &	0.397 &	0.403 &	\uline{0.389} &	\uline{0.399}   \\
\midrule
\multirow{2}{*}{Electricity} & IID   & \textbf{0.177} &	\textbf{0.272} &	0.210 &	0.298 &	0.184 &	\uline{0.273} &	0.203 &	0.294 &	\uline{0.180} &	0.277 &	0.205 &	0.292    \\
& non-IID  & \uline{0.232} &	\uline{0.318} &	0.239 &	0.332 &	0.241 &	0.324 &	\textbf{0.226} &	\textbf{0.313} &	0.298 &	0.381 &	0.233 &	0.318  \\
\midrule
\multirow{2}{*}{Traffic} & IID   & \textbf{0.480} &	\textbf{0.317} &	0.631 &	0.390 &	\uline{0.490} &	\uline{0.329} &	0.517 &	0.335 &	0.514 &	0.337 &	0.519 &	0.330    \\
& non-IID  & \textbf{0.612} &	\textbf{0.372} &	0.677 &	0.419 &	0.663 &	0.420 &	\uline{0.636} &	\uline{0.397} &	0.651 &	0.410 &	0.640 &	0.401  \\
\midrule
\multirow{2}{*}{Exchange} & IID   & \uline{0.372} &	\textbf{0.407} &	\textbf{0.355} &	0.411 &	0.430 &	0.450 &	0.408 &	0.424 &	0.425 &	0.436 &	0.378 &	\uline{0.410}    \\
& non-IID  & \textbf{0.371} &	\textbf{0.410} &	0.390 &	0.448 &	\uline{0.371} &	\uline{0.413} &	0.387 &	0.422 &	0.528 &	0.445 &	0.388 &	0.421 \\
\midrule
\multirow{2}{*}{Weather} & IID   & \textbf{0.243 }&	\textbf{0.272} &	0.264 &	0.313 &	0.268 &	0.286 &	0.257 &	0.278 &	\uline{0.243} &	\uline{0.272} &	0.258 &	0.279    \\
& non-IID & \textbf{0.261} &	\textbf{0.286} &	\uline{0.263} &	0.311 &	0.274 &	0.291 &	0.275 &	0.295 &	0.267 &	0.289 &	0.268 &	\uline{0.288}   \\
\bottomrule
\end{tabular}
\end{table*}

\subsection{Non-Federated Forecasting Experiments} 
Most of the time series forecasting models are designed for the non-federated learning scenarios. Therefore, comparing PiXTime with baselines on traditional time-series forecasting tasks is crucial for us to establish credibility. We selected the M2M long-term forecasting task, which is widely adopted in non-federated benchmarks. Table~\ref{single} reports the experimental results across five datasets. For each dataset, the reported metrics represent the average performance across four prediction lengths. Notably, the results for the ETT dataset are further averaged across its four sub-datasets, with each sub-dataset evaluated under the same four prediction lengths.

As shown in Table~\ref{single}, PiXTime exhibits a competitive performance on the non-federated M2M forecasting task. PiXTime achieves the best performance in 6 out of 10 evaluation results, significantly better than the baseline models, which achieve a maximum of one best performance each. Furthermore, averaging results across all five datasets, PiXTime outperforms all baseline models in both MSE and MAE metrics. Specifically, PiXTime achieves relative reductions of 15.26\% in MSE and 12.53\% in MAE compared to DLinear; 4.58\% and 3.17\% compared to iTransformer; 5.12\% and 3.45\% compared to PatchTST; 5.39\% and 3.45\% compared to TimeXer, and 5.66\% and 3.17\% compared to FFTS, respectively. These results demonstrate that PiXTime achieves advanced performance even on conventional time-series forecasting tasks under non-federated settings.


\subsection{Federated Forecasting Results}

As surveyed in Section \ref{related_work}, to the best of our knowledge, PiXTime is among the first models to simultaneously address temporal granularity misalignment and variable set inconsistencies in federated forecasting. Since none of the baselines natively handles such structural heterogeneity, we conduct a two-stage evaluation for fairness. First, we benchmark PiXTime against all baselines in structurally homogeneous federated environments to evaluate its foundational performance. Subsequently, through experiments in heterogeneous environments, we investigate whether PiXTime can achieve further performance enhancements compared to its homogeneous counterparts. By synthesizing these two stages of empirical evaluation—demonstrating baseline superiority in homogeneous settings followed by performance gains under heterogeneity—we aim to systematically validate PiXTime's capability and SOTA potential in structurally heterogeneous federated scenarios.


\subsubsection{Homogeneous Results}\label{fed_result}
Table~\ref{fed} presents the evaluation results of PiXTime and the five baseline models under the federated network setting, where the data structure is identical across all nodes. The task remains the widely adopted long-term M2M forecasting. We evaluate the models under both Independent and Identically Distributed (IID) and Non-IID data partitioning settings. In all cases, the local training datasets are disjoint subsets sampled exclusively from the complete training dataset. Under the IID setting, the data is partitioned via a uniform distribution, ensuring similar local dataset sizes and distributions across nodes; corresponding optimization is performed using FedAvg as outlined in Algorithm~\ref{pixtime_fedavg}. Conversely, under the Non-IID setting, the partitioning follows a Dirichlet distribution (with $\alpha=0.5$), deliberately creating varying local dataset sizes and skewed distributions without influencing the internal structure of the data; optimization is consequently conducted via FedProx with a proximal coefficient of 0.1. During evaluation, each node has the complete test dataset. For PiXTime, after federated training, each node retrieves the shared modules from the server, assembles them with its personalized local modules, evaluates its personalized complete model on the complete test dataset, and uploads the metrics. The server then averages these individual metrics to yield the final results. For the other baselines without local modules, they directly obtain the complete global model from the server and run the same evaluation. Consistent with previous settings, the metrics in Table~\ref{fed} are averaged across four prediction lengths.

As shown in Table~\ref{fed}, PiXTime achieves the best performance on 17 of the 20 evaluated metrics and the second-best on the remaining three. When averaging the results across all five datasets, PiXTime exhibits a dominant advantage over all baselines under both federated settings. Specifically, under the IID setting, PiXTime achieves relative MSE/MAE reductions of 13.68\% / 10.78\% against DLinear, 6.81\% / 4.88\% against iTransformer, 7.60\% / 4.61\% against PatchTST, 5.74\% / 3.77\% against TimeXer, and 6.01\% / 3.21\% against the FL-specific FFTS. Similarly, under the challenging Non-IID setting, PiXTime maintains its superiority, delivering relative MSE/MAE reductions of 7.92\% / 8.92\% (DLinear), 4.61\% / 3.77\% (iTransformer), 4.61\% / 3.25\% (PatchTST), 13.08\% / 7.51\% (TimeXer), and 2.87\% / 2.19\% (FFTS). These empirical results confirm that PiXTime not only establishes SOTA performance for time-series forecasting in federated learning scenarios but also demonstrates superior adaptability and robustness to complex data distributions across nodes.




\begin{table*}[t]
\centering
\caption{Effectiveness study of PiXTime in federated networks with multiple time granularities.}
\label{abl_freq}
\begin{tabular}{c|c|ccc|ccc|ccc|ccc|ccc}
\toprule
\multicolumn{2}{c}{Data} & \multicolumn{3}{c}{ETT-1}& \multicolumn{3}{c}{ETT-2} & \multicolumn{3}{c}{Electricity} & \multicolumn{3}{c}{Traffic} & \multicolumn{3}{c}{Weather}\\
\cmidrule(lr){1-2} \cmidrule(lr){3-5} \cmidrule(lr){6-8} \cmidrule(lr){9-11} \cmidrule(lr){12-14} \cmidrule(lr){15-17}
\multicolumn{2}{c}{Freq}  & mix & h   & m   & mix & h   & m & mix & syn   & ori & mix & syn   & ori & mix & syn   & ori   \\
\midrule
\multirow{2}{*}{Metric} &MSE     & \textbf{0.051} & 0.092 & 0.055 & \textbf{0.121} & 0.224 & 0.129 & \textbf{0.330} & 0.499 & 0.338 & \textbf{0.156} & 0.196 & 0.163 & \textbf{0.001} & 0.002 & 0.001 \\
&MAE     & \textbf{0.170} & 0.233 & 0.175 & \textbf{0.260} & 0.371 & 0.265 & \textbf{0.408} & 0.519 & 0.419 & \textbf{0.239} & 0.266 & 0.245 & \textbf{0.026} & 0.041 & 0.030\\
\bottomrule
\end{tabular}
\end{table*}

\subsubsection{Heterogeneous Results}For multiple temporal granularities, as previously stated, PiXTime's personalized local modules are designed to resolve this structural heterogeneity across nodes. By aligning data from diverse temporal granularities into a unified dimensional space for the shared modules, PiXTime leverages the continuous communication of these shared parameters to achieve effective knowledge transfer across multiple granularities. This multi-granular alignment ultimately enhances the overall forecasting capability of the model across the federated network. To validate the effectiveness of the personalized local modules, we conduct experiments on four of the datasets. Among the evaluated benchmarks, ETT is the only dataset that natively provides multiple temporal granularities (containing both minute-level and hour-level sampling versions for each of its two sub-datasets). For the remaining datasets, we construct coarse-grained synthetic data (``syn'') by applying interval-based resampling to the original (``ori'') sequences based on their respective temporal periodicities. Under an IID data distribution, Table~\ref{abl_freq} reports the average performance of PiXTime on these datasets across four different prediction lengths.

For the frequency setting (``Freq'') in Table~\ref{abl_freq}, taking ETT-1 as an example, the columns ``h'' and ``m'' report the average performance of PiXTime on the M2U forecasting task within a homogeneous federated network trained exclusively on ETT-h1 and ETT-m1, respectively. The evaluation for these two homogeneous settings follows the method in Section \ref{fed_result}. Conversely, the ``mix'' column represents a structurally heterogeneous federated scenario where half of the nodes utilize ETT-h1, while the remaining half employ ETT-m1. To guarantee strict alignment across nodes based on identical physical time intervals despite their divergent sampling rates, the prediction lengths for the ETT-h1 nodes are proportionally adjusted to $\{24, 48, 84, 180\}$. These lengths correspond to the exact temporal spans covered by the prediction lengths defined for ETT-m1; all other time-related hyperparameters are consistently adapted in the same manner. The evaluation for the ``mix'' setting basically aligns with Section \ref{fed_result}, with one crucial distinction: to ensure the execution of the personalized local modules, each node evaluates its complete model exclusively on the full test set corresponding to its specific temporal granularity. The experiments for all other datasets follow similar configurations.

As shown in Table 4, PiXTime trained within the multi-granular ``mix'' federated network comprehensively outperforms its counterparts trained in the structurally homogeneous ``h''/``syn'' and ``m''/``ori'' networks. Notably, this superiority is evident not only when comparing the ``mix'' setting against the coarse-grained networks, but also against the fine-grained networks. The only exception occurs in the MSE of the Weather, which results in a tie because the error has already reached the minimal statistical precision (0.001). Specifically, compared to the coarse-grained and fine-grained baselines respectively, the ``mix'' setting achieves relative reductions in MSE and MAE across all benchmarks: (44.57\%, 27.04\%) and (7.27\%, 2.86\%) on ETT-1; (45.98\%, 29.92\%) and (6.20\%, 1.89\%) on ETT-2; (33.86\%, 21.38\%) and (2.36\%, 2.62\%) on Electricity; (20.40\%, 10.15\%) and (4.29\%, 1.73\%) on Traffic; (50.00\%, 36.58\%) and (-, 13.33\%) on Weather. Synthesizing these observations, it is evident that PiXTime's personalized local modules, built upon the parameter-decoupling baseline, achieve positive transfer across nodes with diverse temporal granularities, yielding stronger performance than training in a single-granularity network.


\begin{figure}[t]
    \centering
    \subfloat[MSE in Electricity]{%
        \includegraphics[width=0.49\linewidth]{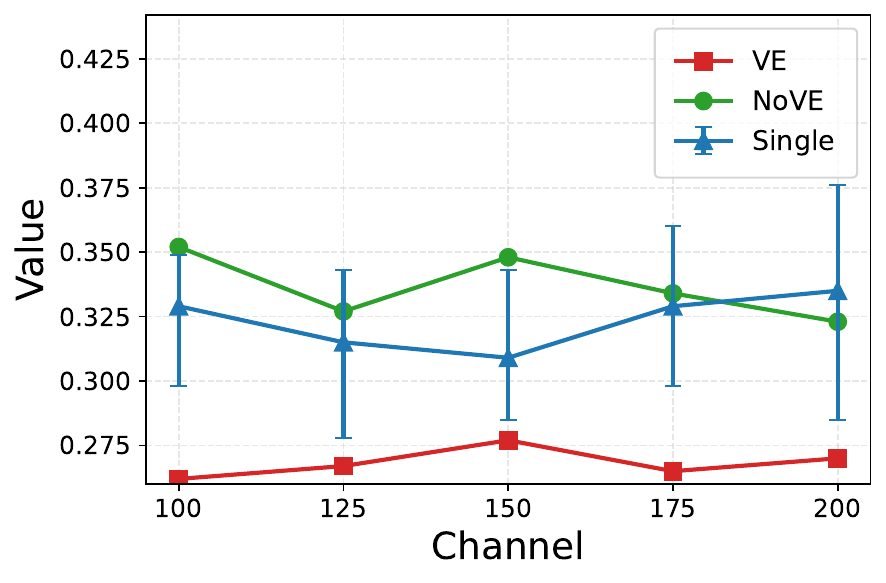}%
    }~
        \subfloat[MAE in Electricity]{%
        \includegraphics[width=0.49\linewidth]{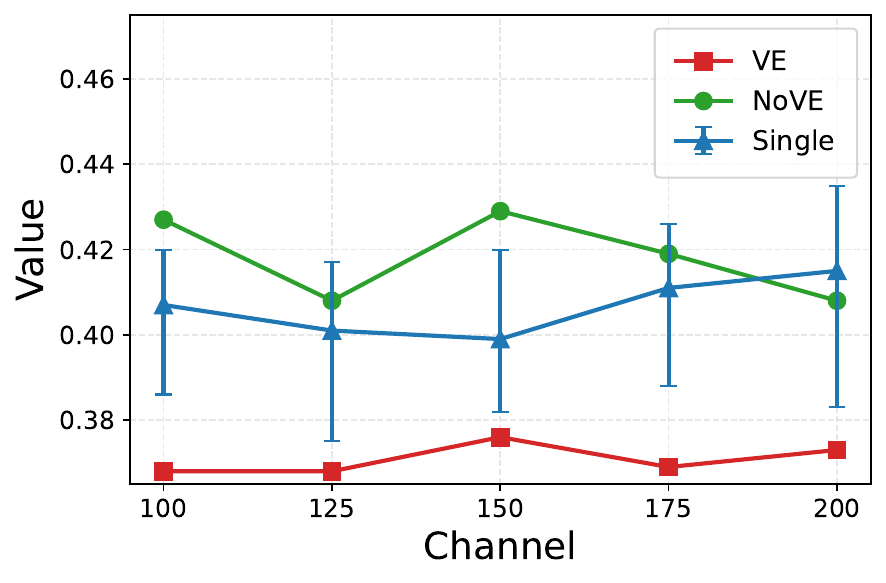}%
    }
    
    \subfloat[MSE in Traffic]{%
        \includegraphics[width=0.49\linewidth]{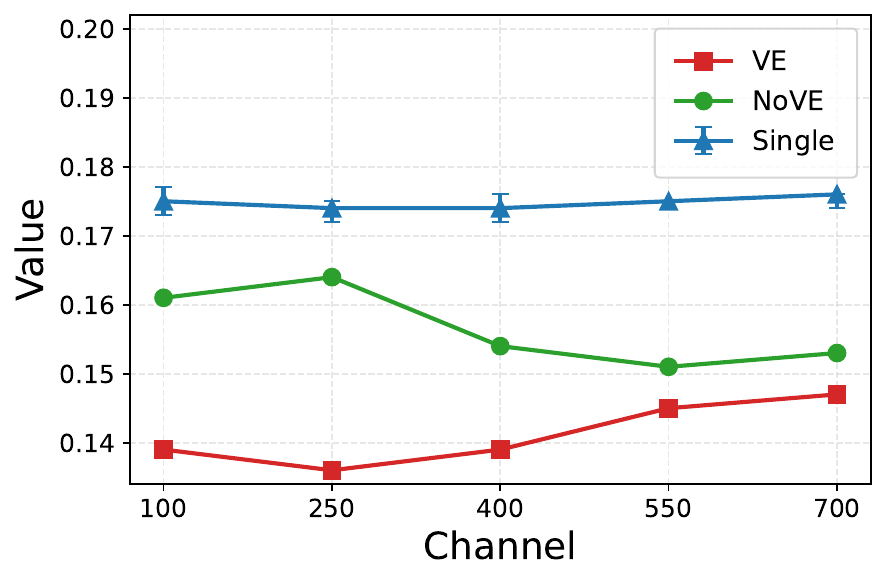}%
    }~
        \subfloat[MAE in Traffic]{%
        \includegraphics[width=0.49\linewidth]{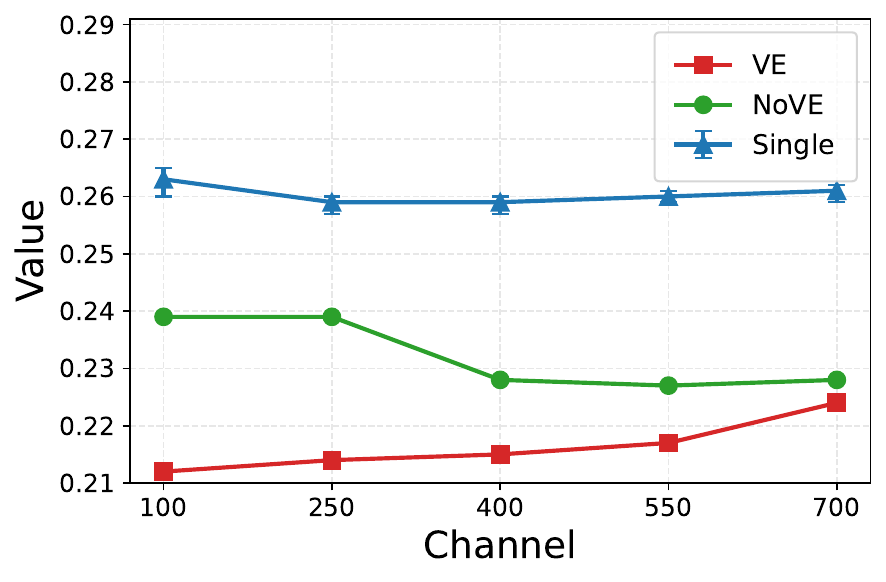}%
    }
    \caption{Effectiveness study of the VE Table in federated networks with heterogeneous auxiliary variable sets.}
    \label{abl_ve}
    \Description{None}
\end{figure}

For diverse variable sets, PiXTime employs the VE Table and its synchronization mechanism to identify variable categories across nodes, enabling cross-node transfer of processing capabilities for identical auxiliary variables and thus improving forecasting performance in federated networks with heterogeneous auxiliary variable sets. To evaluate its effectiveness, we conduct experiments on the Electricity and Traffic datasets, which contain hundreds of variables (321 and 862, respectively), posing a significant challenge for variable representation extraction and transfer across nodes.


Fig.~\ref{abl_ve} compares the performance of PiXTime optimized via federated learning, PiXTime trained independently on individual nodes, and an ablated PiXTime (without the VE Table) optimized via federated learning, all under heterogeneous auxiliary variable sets. Taking the Electricity dataset as an example, each node randomly samples a subset of \{100, 125, 150, 175, 200\} variables to construct its local auxiliary set. In the figure, the ``Single'' and ``VE'' markers show the results of the original PiXTime model, whereas ``NoVE'' illustrates the ablated version without the VE Table. The ``Single'' markers represent the performance of independent training at each node without communication, where the points denote the average metrics across all nodes, and the error bars indicate the maximum and minimum metrics among them. Conversely, the ``VE'' and ``NoVE'' markers represent the performance achieved via collaborative federated learning. Through the training, the data distribution is IID across nodes. Regarding the evaluation, because the task is M2U forecasting, the target variable is identical across all nodes within a setting, despite the heterogeneity in their auxiliary variable sets. Consequently, models under all three settings can be fairly evaluated on the same complete test set. Specifically, under the ``Single'' setting, each node is evaluated independently, with individual node metrics reported in Fig.~\ref{abl_ve}. For the ``VE'' and ``NoVE'' federated settings, the evaluation follows the method detailed in Section \ref{fed_result}. The experiments on Traffic follow the same configurations.


By comparing the ``VE'' and ``Single'' results in Fig.~\ref{abl_ve}, it is evident that under highly heterogeneous variable configurations across nodes, the PiXTime model optimized via federated learning significantly outperforms even the best-performing individual models trained independently. Furthermore, comparing the ``VE'' and ``NoVE'' results reveals the critical role of the VE Table. The integration of the VE Table yields substantial performance gains for PiXTime when operating in networks with highly diverse auxiliary variable sets. Especially on the Electricity dataset, without the VE Table, the performance of the model shown by ``NoVE'' is even lower than the average performance of isolated local training. Synthesizing these observations, we conclude that the VE Table is an indispensable module that enables PiXTime to be effectively optimized via federated learning, boosting performance in networks with severe auxiliary variable heterogeneity relative to training these nodes on their own homogeneous datasets.

\subsubsection{Conclusion for Federated Experiments}
In summary, the empirical evaluations validate PiXTime's effectiveness in federated time series forecasting. Initially, the results in Table~\ref{fed} demonstrate that PiXTime outperforms existing baselines in structurally homogeneous federated environments. Furthermore, when deployed in structurally heterogeneous scenarios, PiXTime leverages its personalized local modules and the VE Table to successfully achieve positive knowledge transfer across diverse temporal granularities and heterogeneous auxiliary variable sets. Notably, PiXTime performs better under these heterogeneous configurations than it does under homogeneous settings. Synthesizing these observations, we conclude that PiXTime effectively resolves the structural incompatibility between heterogeneous data and federated aggregation, establishing a SOTA capability for structurally diverse federated time series forecasting.





\subsection{Ablation Study for PiXTime}

PiXTime introduces an abstract token to bridge the representational granularity gap between the Auxiliary Encoder and the Target Decoder. Furthermore, because the abstract token represents a variable-wise aggregation while the patch tokens retain fine-grained temporal patterns, PiXTime's Projection Head exclusively utilizes the patch tokens to generate the final prediction sequence. To validate the necessity of these two design choices, we conduct corresponding ablation studies, with the results reported in Table~\ref{abl_ph}.

\begin{table}[t]
\centering
\caption{Ablation study on the patch-exclusive projection head and the abstract token in PiXTime.}
\label{abl_ph}
\resizebox{\columnwidth}{!}{
\begin{tabular}{c|c|cccccc}
\toprule
\multicolumn{2}{c}{Data} & ETT & Electricity & Traffic & Exchange & Weather \\
\midrule
\multirow{2}{*}{Ori}  & MSE & 0.377  &  0.186  & 0.486 & 0.371 & 0.245 \\
& MAE & 0.392 & 0.282  & 0.319 &  0.411 &  0.273\\

\midrule
\multirow{2}{*}{PH}  & MSE &  0.381 &  0.192  & 0.489 &   0.377 & 0.248 \\
& MAE &  0.397 &  0.286  & 0.323 &   0.412 & 0.277 \\
\midrule
\multirow{2}{*}{Abs}  & MSE &  0.382 &  0.204  & OOM &   0.371 & 0.258 \\
& MAE &  0.398 &  0.297  & OOM &   0.411 & 0.282 \\
\bottomrule
\end{tabular}
}
\end{table}

In Table~\ref{abl_ph}, ``Ori'' denotes the original PiXTime architecture, while these results are directly adopted from Table \ref{single}, as the experimental configurations are identical. ``PH'' represents an ablated variant where the abstract token is concatenated with the patch tokens and jointly fed into the Projection Head. Consequently, the original projection formulation in Eq.~\eqref{PH} is modified to:
\begin{equation}\label{abl_PH}
\textbf{x}_i^{pre} = Projection\left ([\textbf{a}_{i,L}, \textbf{P}_{i,L} ]\right ),
\end{equation}
with all other components remaining unchanged. ``Abs'' follows the same variant described in the experiments of Table~\ref{tab:pixtime_specs}. The experimental environments, datasets, and hyperparameters for all three models are kept strictly identical to ensure a fair comparison.

Comparing the experimental results of ``Ori'' and ``PH'' in Table~\ref{abl_ph}, it is evident that ``Ori'' outperforms ``PH'' in all of the 10 evaluated metrics. Similarly, comparing ``Ori'' against the ``Abs'' variant, ``Ori'' exhibits clear superiority by achieving better performance in 8 out of the 10 metrics, while maintaining the same in the remaining 2. More crucially, beyond predictive accuracy, we observed that the ``Abs'' variant triggered an Out-of-Memory (OOM) error during training on the Traffic dataset under identical hardware (on a NVIDIA Tesla V100-SXM2 32G GPU) and parameter configurations. This phenomenon is consistent with the larger maximum memory footprint and computational overhead reported for "Abs" in Table 1, which are caused by the excessively high number of cross-attention queries in "Abs". This results in an uncontrollable memory overhead when processing the Traffic dataset, which involves up to 862 target variables, directly causing the OOM failure. In summary, the comprehensive results in Table~\ref{abl_ph} validate the efficiency introduced by the abstract token and the Projection Head strategy in PiXTime.




%% file: sections/conclusion.tex
\section{Conclusion}


In this paper, we addressed a critical bottleneck in federated time series forecasting: achieving efficient collaborative training across nodes with structurally heterogeneous data. To resolve this, we proposed PiXTime, a parameter-decoupled forecasting model and its corresponding federated framework, specifically designed to tackle the two primary structural heterogeneities: diverse temporal granularities and inconsistent variable sets. Extensive empirical evaluations validated our approach. PiXTime maintains highly competitive performance in single-node tasks. In federated scenarios, it established a new SOTA performance baseline under structurally homogeneous settings. Furthermore, controlled experiments demonstrated PiXTime's capability to achieve positive knowledge transfer across heterogeneous settings, yielding stronger performance than training in homogeneous networks. Based on these progressive results, we conclude that PiXTime bridges the gap of heterogeneous data structures across distributed nodes, providing an effective and practical solution for real-world tasks.


%% file: sample-base.bib
@String{Computing = "Computing" }

@String{Computer = "{IEEE} Computer" }

@String{Springer = "Springer-Verlag" }

@ArtifactSoftware{R,
    title = {R: A Language and Environment for Statistical Computing},
    author = {{R Core Team}},
    organization = {R Foundation for Statistical Computing},
    address = {Vienna, Austria},
    year = {2019},
    url = {https://www.R-project.org/},
}


%% file: sections/cite.bib
@inproceedings{Exchange,
  title={Modeling Long- and Short-Term Temporal Patterns with Deep Neural Networks},
  author={Lai, Guokun and Chang, Wei-Cheng and Yang, Yiming and Liu, Hanxiao},
  booktitle = {The 41st International ACM SIGIR Conference on Research and Development in Information Retrieval (SIGIR '18)},
  pages={95--104},
  year={2018},
  publisher={ACM},
  address = {New York, NY, USA}
}

@inproceedings{Code,
  title={TimesNet: Temporal 2D-Variation Modeling for General Time Series Analysis},
  author={Wu, Haixu and Hu, Tengge and Liu, Yong and Zhou, Hang and Wang, Jianmin and Long, Mingsheng},
  booktitle={Proceedings of the International Conference on Learning Representations (ICLR '23)},
  year={2023},
  publisher={OpenReview.net},
  address={Online}
}

@inproceedings{Informer,
  title={Informer: Beyond Efficient Transformer for Long Sequence Time-Series Forecasting},
  author={Zhou, Haoyi and Zhang, Shanghang and Peng, Jieqi and Zhang, Shuai and Li, Jianxin and Xiong, Hui and Zhang, Wancai},
  booktitle={Proceedings of the AAAI Conference on Artificial Intelligence (AAAI '21)},
  volume={35},
  pages={11106--11115},
  year={2021},
  publisher={AAAI},
  address = {CA, USA}
}

@inproceedings{Autoformer,
  title={Autoformer: Decomposition Transformers with Auto-Correlation for Long-Term Series Forecasting},
  author={Wu, Haixu and Xu, Jiehui and Wang, Jianmin and Long, Mingsheng},
  booktitle={Advances in Neural Information Processing Systems (NeurIPS '21)},
  volume={34},
  pages={22419--22430},
  year={2021},
  publisher = {Curran Associates, Inc.},
  address={NY, USA}
}

@inproceedings{DLinear,
  title={Are Transformers Effective for Time Series Forecasting?},
  author={Zeng, Ailing and Chen, Muxi and Zhang, Lei and Xu, Qiang},
  booktitle={Proceedings of the AAAI Conference on Artificial Intelligence (AAAI '23)},
  volume={37},
  pages={11121--11128},
  year={2023},
  publisher={AAAI},
  address = {CA, USA}
}

@inproceedings{PatchTST,
  title={A Time Series is Worth 64 Words: Long-Term Forecasting with Transformers},
  author={Nie, Yuqi and Nguyen, Nam H and Sinthong, Phanwadee and Kalagnanam, Jayant},
  booktitle={Proceedings of the International Conference on Learning Representations (ICLR '23)},
  year={2023},
  publisher={OpenReview.net},
  address={Online}
}

@inproceedings{TimeXer,
  title={TimeXer: Empowering Transformers for Time Series Forecasting with Exogenous Variables},
  author={Wang, Yuxuan and Wu, Haixu and Dong, Jiaxiang and Qin, Guo and Zhang, Haoran and Liu, Yong and Qiu, Yunzhong and Wang, Jianmin and Long, Mingsheng},
  booktitle={Advances in Neural Information Processing Systems (NeurIPS '24)},
  volume={37},
  pages={469--498},
  year={2024},
  publisher = {Curran Associates, Inc.},
  address={NY, USA}
}

@inproceedings{iTrans,
  title={i{T}ransformer: {I}nverted Transformers Are Effective for Time Series Forecasting},
  author={Liu, Yong and Hu, Tengge and Zhang, Haoran and Wu, Haixu and Wang, Shiyu and Ma, Lintao and Long, Mingsheng},
  booktitle={Proceedings of the International Conference on Learning Representations (ICLR '24)},
  pages = {11116--11140},
  year={2024},
  publisher={OpenReview.net},
  address={Online}
}

@inproceedings{EAPformer,
  title={{EAP}former: {E}ntropy-Aware Patch Transformer for Multivariate Long-Term Time Series Forecasting},
  author={Ling, Jiahao and Yang, Xuan and Gong, Shimin and Gu, Bo},
  booktitle={Proceedings of the 34th ACM International Conference on Information and Knowledge Management (CIKM '25)},
  pages={1850--1860},
  year={2025},
  publisher =    {ACM},
  address = {New York, NY, USA}
}

@inproceedings{AdaPatch,
  title={Ada{P}atch: {A}daptive Patch-Level Modeling for Non-Stationary Time Series Forecasting},
  author={Liu, Kun and Duan, Zhongjie and Chen, Cen and Wang, Yanhao and Cheng, Dawei and Liang, Yuqi},
  booktitle={Proceedings of the 34th ACM International Conference on Information and Knowledge Management (CIKM '25)},
  pages={1882--1891},
  year={2025},
  publisher =    {ACM},  
  address = {New York, NY, USA}
}

@inproceedings{FinCast,
  title={Fin{C}ast: {A} Foundation Model for Financial Time-Series Forecasting},
  author={Zhu, Zhuohang and Chen, Haodong and Qu, Qiang and Chung, Vera},
  booktitle={Proceedings of the 34th ACM International Conference on Information and Knowledge Management (CIKM '25)},
  pages={4539--4549},
  year={2025},
  publisher =    {ACM},
  address = {New York, NY, USA}
}

@inproceedings{HyperIMTS,
  title = 	 {{H}yper{IMTS}: {H}ypergraph Neural Network for Irregular Multivariate Time Series Forecasting},
  author =       {Li, Boyuan and Luo, Yicheng and Liu, Zhen and Zheng, Junhao and Lv, Jianming and Ma, Qianli},
  booktitle = 	 {Proceedings of the 42nd International Conference on Machine Learning (ICML '25)},
  pages = 	 {35502--35518},
  year = 	 {2025},
  volume = 	 {267},
  series = 	 {Proceedings of Machine Learning Research},
  month = 	 {13--19 Jul},
  publisher =    {PMLR},
  address={Online}
}

@inproceedings{TimeFM,
  title = 	 {A Decoder-Only Foundation Model for Time-Series Forecasting},
  author =       {Das, Abhimanyu and Kong, Weihao and Sen, Rajat and Zhou, Yichen},
  booktitle = 	 {Proceedings of the 41st International Conference on Machine Learning (ICML '24)},
  pages = 	 {10148--10167},
  year = 	 {2024},
  volume = 	 {235},
  series = 	 {Proceedings of Machine Learning Research},
  month = 	 {21--27 Jul},
  publisher =    {PMLR},
  address={Online}
}

@inproceedings{FFTS,
  title={Federated Foundation Models on Heterogeneous Time Series},
  author={Chen, Shengchao and Long, Guodong and Jiang, Jing and Zhang, Chengqi},
  booktitle={Proceedings of the AAAI Conference on Artificial Intelligence (AAAI '25)},
  volume={39},
  pages={15839--15847},
  year={2025},
  publisher={AAAI},
  address = {CA, USA}
}

@inproceedings{TimeFFM,
  title={Time-{FFM}: {T}owards LM-Empowered Federated Foundation Model for Time Series Forecasting},
  author={Liu, Qingxiang and Liu, Xu and Liu, Chenghao and Wen, Qingsong and Liang, Yuxuan},
  booktitle={Advances in Neural Information Processing Systems (NeurIPS '24)},
  volume={37},
  pages={94512--94538},
  year={2024},
  publisher = {Curran Associates, Inc.},
  address={NY, USA}
}

@article{ProtoPFL,
  title={Proto{PFL}: {E}nhancing Privacy and Accuracy in Federated Time-Series Forecasting with Heterogeneous Data},
  author={Chen, Shan and Chen, Ke and Zhao, Xinzhe and Yu, Minming and Zhang, Tao and Wang, Wenqi},
  journal={Computer Standards \& Interfaces},
  volume = {98},
  pages={104161},
  year={2026},
  publisher={Elsevier}
}

@inproceedings{WeatherLLM,
  title={Personalized Adapter for Large Meteorology Model on Devices: {T}owards Weather Foundation Models},
  author={Chen, Shengchao and Long, Guodong and Jiang, Jing and Zhang, Chengqi},
  booktitle={Advances in Neural Information Processing Systems (NeurIPS '24)},
  volume={37},
  pages={84897--84943},
  year={2024},
  publisher = {Curran Associates, Inc.},
  address={NY, USA}
}

@inproceedings{FedRMamba,
  title={FedRMamba: Federated Residual Mamba for Multivariate Time-Series Forecasting},
  author={Hu, Zhiwei and Zhang, Liang and Zhu, Guangxu},
  booktitle={Proceedings of the ACM Web Conference 2026 (WWW '26)},
  pages={7610--7620},
  year={2026},
  publisher = {ACM},
  address = {New York, NY, USA}
}

@inproceedings{Tang,
  title={Optimal Look-back Horizon for Time Series Forecasting in Federated Learning},
  author={Tang, Dahao and Yang, Nan and Li, Yanli and Zhu, Zhiyu and Jin, Zhibo and Yuan, Dong},
  booktitle={Proceedings of the AAAI Conference on Artificial Intelligence (AAAI '26)},
  volume={40},
  pages={25823--25830},
  year={2026},
  publisher={AAAI},
  address = {CA, USA}
}

@inproceedings{Ali,
  author={Ali, Mahad and Lisle, Curtis and Moore, Patrick W. and Barkouki, Tammer and Kirkwood, Brian J. and Brattain, Laura J.},
  booktitle={47th Annual International Conference of the IEEE Engineering in Medicine and Biology Society (EMBC '25)}, 
  title={Fine-Tuning Foundation Models with Federated Learning for Privacy Preserving Medical Time Series Forecasting}, 
  year={2025},
  pages={1--7},
  publisher={IEEE},
  address={NY, USA}
}

@inproceedings{FeDal,
  title={FeDal: Federated Dataset Learning for General Time Series Foundation Models},
  author={Chen, Shengchao and Long, Guodong and Blumenstein, Michael and Jiang, Jing},
  booktitle={Proceedings of the 14th International Conference on Learning Representations (ICLR '26)},
  year={2026},
  publisher={OpenReview.net},
  address={Online}
}

@article{FedTREND,
  title={Tackling Data Heterogeneity in Federated Time Series Forecasting},
  author={Yuan, Wei and Yang, Chaoqun and Zhao, Xiangyu and Nguyen, Quoc Viet Hung and Cao, Yang and He, Tieke and Yin, Hongzhi},
  journal={Science China Information Sciences},
  volume={69},
  pages={152102},
  year={2026},
  publisher={Springer}
}

@inproceedings{FedTime,
  title={A Federated Large Language Model for Long-Term Time Series Forecasting},
  author={Abdel-Sater, Raed and Ben Hamza, A.},
  booktitle={Proceedings of the 27th European Conference on Artificial Intelligence (ECAI '24)},
  pages={2452--2459},
  year={2024},
  publisher={SAGE Publications Ltd.},
  address={London, UK}
}

@inproceedings{CrossSiloFed,
  title={Personalized Cross-Silo Federated Learning on {N}on-{IID} Data},
  author={Huang, Yutao and Chu, Lingyang and Zhou, Zirui and Wang, Lanjun and Liu, Jiangchuan and Pei, Jian and Zhang, Yong},
  booktitle={Proceedings of the AAAI Conference on Artificial Intelligence (AAAI '21)},
  volume={35},
  pages={7865--7873},
  year={2021},
  publisher={AAAI},
  address = {CA, USA}
}

@inproceedings{PerCrossSiloFed,
  title={Fed{APEN}: {P}ersonalized Cross-Silo Federated Learning with Adaptability to Statistical Heterogeneity},
  author={Qin, Zhen and Deng, Shuiguang and Zhao, Mingyu and Yan, Xueqiang},
  booktitle={Proceedings of the 29th ACM SIGKDD Conference. on Knowledge Discovery and Data Mining (KDD '23)},
  pages={1954--1964},
  year={2023},
  publisher = {ACM},
  address = {New York, NY, USA}
}

@inproceedings{SCAFFOLD,
  title={{SCAFFOLD}: {S}tochastic Controlled Averaging for Federated Learning},
  author={Karimireddy, Sai Praneeth and Kale, Satyen and Mohri, Mehryar and Reddi, Sashank and Stich, Sebastian and Suresh, Ananda Theertha},
  booktitle={Proceedings of the 37th International Conference on Machine Learning (ICML '20)},
  pages={5132--5143},
  year={2020},
  publisher = {PMLR},
  address={Online}
}

@inproceedings{FedAVG,
  title={Communication-Efficient Learning of Deep Networks from Decentralized Data},
  author={McMahan, Brendan and Moore, Eider and Ramage, Daniel and Hampson, Seth and y Arcas, Blaise Aguera},
  booktitle={Proceedings of the 20th International Conference on Artificial Intelligence and Statistics (AISTATS '17)},
  pages={1273--1282},
  year={2017},  
  publisher = {PMLR},
  address={Online}
}

@inproceedings{FedProx,
  title={Federated Optimization in Heterogeneous Networks},
  author={Li, Tian and Sahu, Anit Kumar and Zaheer, Manzil and Sanjabi, Maziar and Talwalkar, Ameet and Smith, Virginia},
  booktitle={Proceedings of Machine Learning and Systems (MLSys '20)},
  pages={429--450},
  volume={2},
  year={2020}
}

@inproceedings{pFedMe,
  title={Personalized Federated Learning with Moreau Envelopes},
  author={T Dinh, Canh and Tran, Nguyen and Nguyen, Josh},
  booktitle={Advances in Neural Information Processing Systems (NeurIPS '20)},
  volume={33},
  pages={21394--21405},
  year={2020},
  publisher = {Curran Associates, Inc.},
  address={NY, USA}
}

@article{VRSGD,
  title={Accelerating Local {SGD} for {N}on-{IID} Data Using Variance Reduction},
  author={Liang, Xianfeng and Shen, Shuheng and Chen, Enhong and Liu, Jinchang and Liu, Qi and Cheng, Yifei and Pan, Zhen},
  journal={Frontiers of Computer Science},
  volume={17},
  number={2},
  pages={172311},
  year={2023},
  publisher={Springer}
}

@article{AWPS,
  title={Adaptive Weighting Push-SUM for Decentralized Optimization with Statistical Diversity},
  author={Zhou, Yiming and Cheng, Yifei and Xu, Linli and Chen, Enhong},
  journal={IEEE Transactions on Control of Network Systems},
  year={2025},
  volume={12},
  number={3},
  pages={2337-2349},
  publisher={IEEE},
  address={NY, USA}
}

@article{PS,
  author={Nedić, Angelia and Olshevsky, Alex},
  journal={IEEE Transactions on Automatic Control}, 
  title={Distributed Optimization Over Time-Varying Directed Graphs}, 
  year={2015},
  volume={60},
  number={3},
  pages={601-615},
  publisher={IEEE},
  address={NY, USA}
}

@article{FedPer,
  title={Federated Learning with Personalization Layers},
  author={Arivazhagan, Manoj Ghuhan and Aggarwal, Vinay and Singh, Aaditya Kumar and Choudhary, Sunav},
  journal={arXiv preprint arXiv:1912.00818},
  year={2019}
}

@article{LGFEDAVG,
  title={Think Locally, Act Globally: Federated Learning with Local and Global Representations},
  author={Liang, Paul Pu and Liu, Terrance and Ziyin, Liu and Allen, Nicholas B and Auerbach, Randy P and Brent, David and Salakhutdinov, Ruslan and Morency, Louis-Philippe},
  journal={arXiv preprint arXiv:2001.01523},
  year={2020}
}

@inproceedings{FedRep,
  title={Exploiting Shared Representations for Personalized Federated Learning},
  author={Collins, Liam and Hassani, Hamed and Mokhtari, Aryan and Shakkottai, Sanjay},
  booktitle={Proceedings of the 38th International Conference on Machine Learning (ICML '21)},
  pages={2089--2099},
  year={2021},
  publisher={PMLR},
  address={Online}
}

@inproceedings{ModelDecoupNonConvex,
  title={Federated learning with partial model personalization},
  author={Pillutla, Krishna and Malik, Kshitiz and Mohamed, Abdel-Rahman and Rabbat, Mike and Sanjabi, Maziar and Xiao, Lin},
  booktitle={Proceedings of the 38th International Conference on Machine Learning (ICML '22)},
  pages={17716--17758},
  year={2022},
  publisher={PMLR},
  address={Online}
}

@article{FedBone,
  title={Fed{B}one: {T}owards Large-Scale Federated Multi-Task Learning},
  author={Chen, Yiqiang and Zhang, Teng and Jiang, Xinlong and Chen, Qian and Gao, Chenlong and Huang, Wuliang},
  journal={Journal of Computer Science and Technology},
  volume={39},
  number={5},
  pages={1040--1057},
  year={2024},
  publisher={Springer}
}

@inproceedings{DFedPGP,
  title={Decentralized Directed Collaboration for Personalized Federated Learning},
  author={Liu, Yingqi and Shi, Yifan and Li, Qinglun and Wu, Baoyuan and Wang, Xueqian and Shen, Li},
  booktitle={Proceedings of the IEEE/CVF Conference on Computer Vision and Pattern Recognition (CVPR '24)},
  pages={23168--23178},
  year={2024},
  publisher={IEEE},
  address={NY, USA}
}

@inproceedings{FedBABU,
  title={FedBABU: Towards Enhanced Representation for Federated Image Classification},
  author={Oh, Jaehoon and Kim, Sangmook and Yun, Seyoung},
  booktitle={Proceedings of the 10th International Conference on Learning Representations (ICLR '22)},
  year={2022},
  publisher={OpenReview.net},
  address={Online}
}

@inproceedings{FedRoD,
  title={On Bridging Generic and Personalized Federated Learning for Image Classification},
  author={Chen, Hongyou and Chao, Weilun},
  booktitle={Proceedings of the 10th International Conference on Learning Representations (ICLR '22)},
  year={2022},
  publisher={OpenReview.net},
  address={Online}
}

@inproceedings{FedDecomp,
  title={Decoupling General and Personalized Knowledge in Federated Learning via Additive and Low-Rank Decomposition},
  author={Wu, Xinghao and Liu, Xuefeng and Niu, Jianwei and Wang, Haolin and Tang, Shaojie and Zhu, Guogang and Su, Hao},
  booktitle={Proceedings of the 32nd ACM International Conference on Multimedia (MM '24)},
  pages={7172--7181},
  year={2024},
  publisher={ACM},
  address = {New York, NY, USA}
}
